\pdfoutput=1

\documentclass[11pt]{article}

\usepackage[]{ACL2023}

\usepackage{times}
\usepackage{latexsym}

\usepackage[T1]{fontenc}

\usepackage[utf8]{inputenc}

\usepackage{microtype}

\usepackage{inconsolata}

\usepackage{hyperref}       
\usepackage{soul}

\usepackage{booktabs}       
\usepackage{amsfonts}       
\usepackage{amsmath}
\usepackage{nicefrac}       
\usepackage{microtype}      
\usepackage{xcolor}         
\usepackage{multirow}
\usepackage{caption}
\usepackage{subcaption}
\usepackage{graphicx}
\usepackage{algorithm}
\usepackage{algpseudocode}[1]

\newcommand{\model}{Skill-LM}

\title{Learning Non-linguistic Skills without Sacrificing Linguistic Proficiency}

\author{Mandar Sharma \\
  Virginia Tech \\
  \texttt{mandarsharma@vt.edu} \\\And
  Nikhil Muralidhar \\
  Stevens Institute of Technology \\
  \texttt{nmurali1@stevens.edu} \\\And
  Naren Ramakrishnan \\
  Virginia Tech \\
  \texttt{naren@cs.vt.edu} \\}

\begin{document}
\maketitle
\begin{abstract}
The field of Math-NLP has witnessed significant growth in recent years, motivated by the desire to expand LLM performance to the learning of non-linguistic notions (numerals, and subsequently, arithmetic reasoning). However, non-linguistic skill injection typically comes at a cost for LLMs: it leads to catastrophic forgetting of core linguistic skills, a consequence that often remains unaddressed in the literature. As Math-NLP has been able to create LLMs that can closely approximate the mathematical skills of a grade-schooler or the arithmetic reasoning skills of a calculator, the practicality of these models fail if they concomitantly shed their linguistic capabilities. In this work, we take a closer look into the phenomena of catastrophic forgetting as it pertains to LLMs and subsequently offer a novel framework for non-linguistic skill injection for LLMs based on information-theoretic interventions and skill-specific losses that enable the learning of strict arithmetic reasoning. Our model outperforms the state-of-the-art both on \textit{injected non-linguistic skills} and on  \textit{linguistic knowledge retention}, and does so with a fraction of the non-linguistic training data ($1/4$) and zero additional synthetic linguistic training data. Our pre-trained models and experimentation codebases are hosted online\footnote{\url{https://github.com/Mandar-Sharma/Skill-LM}}.
\end{abstract}

\section{Introduction}
\begin{figure}[ht]
    \centering
    \includegraphics[scale=0.24]{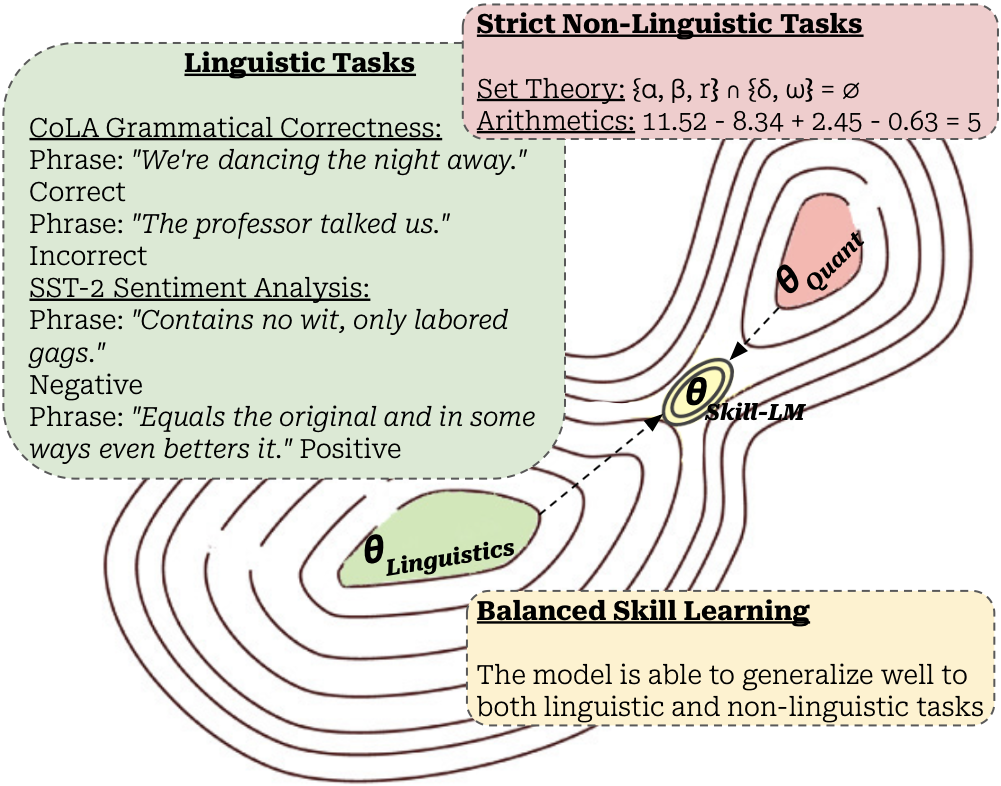}
    \caption{LLMs trained for dissimilar skillsets have different convergence points for their parameters - the parameterization space for an LLM trained for linguistic skills $\theta_{Linguistics}$ lives in the green space while the parameterization space for an LLM trained for quantitative reasoning $\theta_{Quant}$ lives in the red space. The goal of this work is to approximate a locality of parameterization $\theta_{Skill-LM}$ (yellow) where the model reliably learns a non-linguistic skill (quantitative reasoning) without sacrificing its linguistic proficiency.}%
    \label{fig:examples}
\end{figure}
\begin{table*}[t]
\centering
\footnotesize
\begin{tabular}{lcccccccccc}
\hline
Model        & CoLA          & STS-B         & MNLI           & MNLI$_{MM}$        & MRPC           & QNLI           & QQP            & RTE            & SST-2          & WNLI           \\ \hline
BERT         & \textit{0.59} & \textit{0.89} & \textit{83.85} & \textit{84.05} & \textit{86.76} & \textit{90.55} & \textit{90.61} & \textit{65.34} & \textit{91.62} & \textit{56.33} \\
BERT$_{Arith}$ & 0.08          & 0.80          & 32.73          & 32.95          & 70.34          & 50.53          & 70.49          & 47.29          & 88.07          & 56.33          \\ \hline
\end{tabular}
\caption{\textit{LLMs trained for niche non-linguistic skill-sets forget linguistics:} Comparative analysis between the performance of the base BERT model and the same model further trained on an arithmetic reasoning corpus on the set of 9 GLUE tasks for natural language understanding. All tasks except WNLI suffer severe performance degradation as a consequence of continued training on a non-linguistic corpus.}
\label{table:1}
\end{table*}

 Numeracy, involving the comprehension of sizes, magnitudes, and order, is the most prevalent form of \textit{non-linguistic} information embedded in textual corpora~\citep{joram:1995}. Thus, the case for numerically-capable LLMs is rather easy to make: as numerals grant objectivity to language \citep{porter:1996}, numerically-capable language models are key to optimal performance in a host of downstream tasks such as information extraction \citep{madaan:2016}, inference \citep{naik:2018}, and data-to-text generation \cite{sharma:2021,sharma:2022}.
 
\subsection{Re-thinking the Objective of Math-NLP}

\textbf{Progress in Math-NLP:} Several notable publications in the Math-NLP space have made rapid strides in numeracy-tinged language-modeling \citep{thawani:2021} - from investigations of the inherent deficiency of numerical reasoning skills in LLMs induced through unsupervised training, both for numerals that appear in the training corpus \citep{zhang:2020} and OOD (out-of-domain) numerals \citep{wallace:2019, razeghi:2022}, to interventions that strengthen the numerical reasoning skills of these models \citep{spith:2018, jiang:2020, geva:2020}. Further, advances in chain-of-thought prompting in few-shot learning settings \citep{li:2022} and task-specific fine-tuning \citep{lew:2022} have shown significant gains in the capacity for quantitative reasoning in LLMs.

\vspace{0.2cm}
\noindent\textbf{Linguistic evaluation remains important:} As notable as these accomplishments are, the goal remains not to replicate the reasoning capabilities of a grade-schooler or to proxy a calculator, but rather build LLMs that are \textit{empowered} with these skills. As such, an area that often goes unaddressed in the Math-NLP space is how these models perform as \textit{general language modelers}. With the advent and popularity of generative conversational models \cite{chatgpt}, the goal is to have one model capable of a host of skills - not to load separate models for conversation/assistance and reasoning. As depicted in Figure \ref{fig:examples}, whether a model is designed to perform strict non-linguistic tasks or semi-linguistic tasks, it should never come at the cost of core linguistic competency. After all, language models are intended to \textit{model language}.

\subsection{Necessitating the Re-thinking}

\textbf{LLMs injected with non-linguistic skills forgo their linguistic skills:} Consider the task of strict arithmetic reasoning as shown in Figure \ref{fig:examples}, a subset of possible quantitative reasoning tasks. If a base BERT model \citep{bert} is further trained on this non-linguistic task, it suffers significant degradation on 8/9 GLUE tasks \citep{glue} that evaluate the natural language understanding (NLU) capabilties of the model, as showcased in Table \ref{table:1}. This observation has long been known in the deep learning literature as \textit{catastrophic forgetting} \citep{kirkpatrick:2017}, wherein when a model pre-trained on task $A$ is further trained on task $B$, the parameters in the model vital for task $A$ adapt their values to meet the requirements of task $B$. 

\vspace{0.2cm}
\noindent\textbf{LLMs exhibit unconventional forgetting:} What is interesting, based on our findings, is that in the case of LLMs, the forgetting of linguistic skills is not evenly spread - the forgetting is rather \textit{task-specific}. Akin to other neural network applications, the forgetting of linguistic skills may likely be grouped as performance loss over a single task $A$; however, as seen in Table \ref{table:1}, the GLUE tasks suffer various ranges of degradation - the task of finding the referent of a pronoun (WNLI, \citet{wnli}) does not seem to suffer at all, while the grammatical correctness assessment task (CoLA, \citet{cola}) suffers severe degradation. 

As proponents for \textit{skill-empowered} LLMs, we thus make a case for disclosing the performance on general NLU tasks when models are trained for superior performance on niche skill-sets such as non-linguistics, an area left wanting in the Math-NLP front. Because of this task-specific forgetting, quantitative reasoning models trained in a Q\&A fashion may not showcase degradation in similarly modeled downstream tasks such as SQuAD \citep{squad} and DROP \cite{drop} - thus disclosing performance across a range of NLU tasks is crucial.

\begin{figure*}[t]
    \centering
    \includegraphics[scale=0.28]{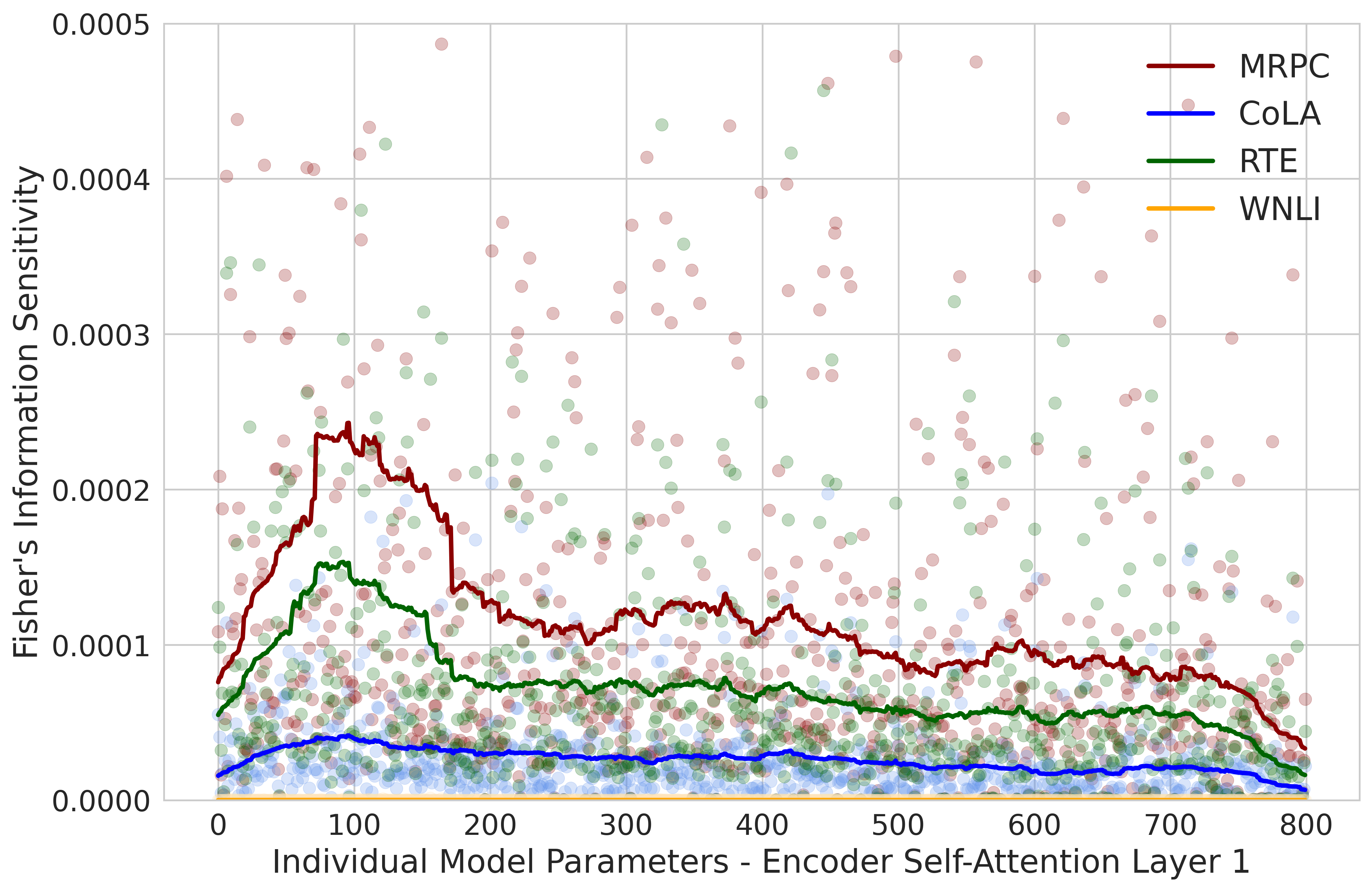}
    \includegraphics[scale=0.28]{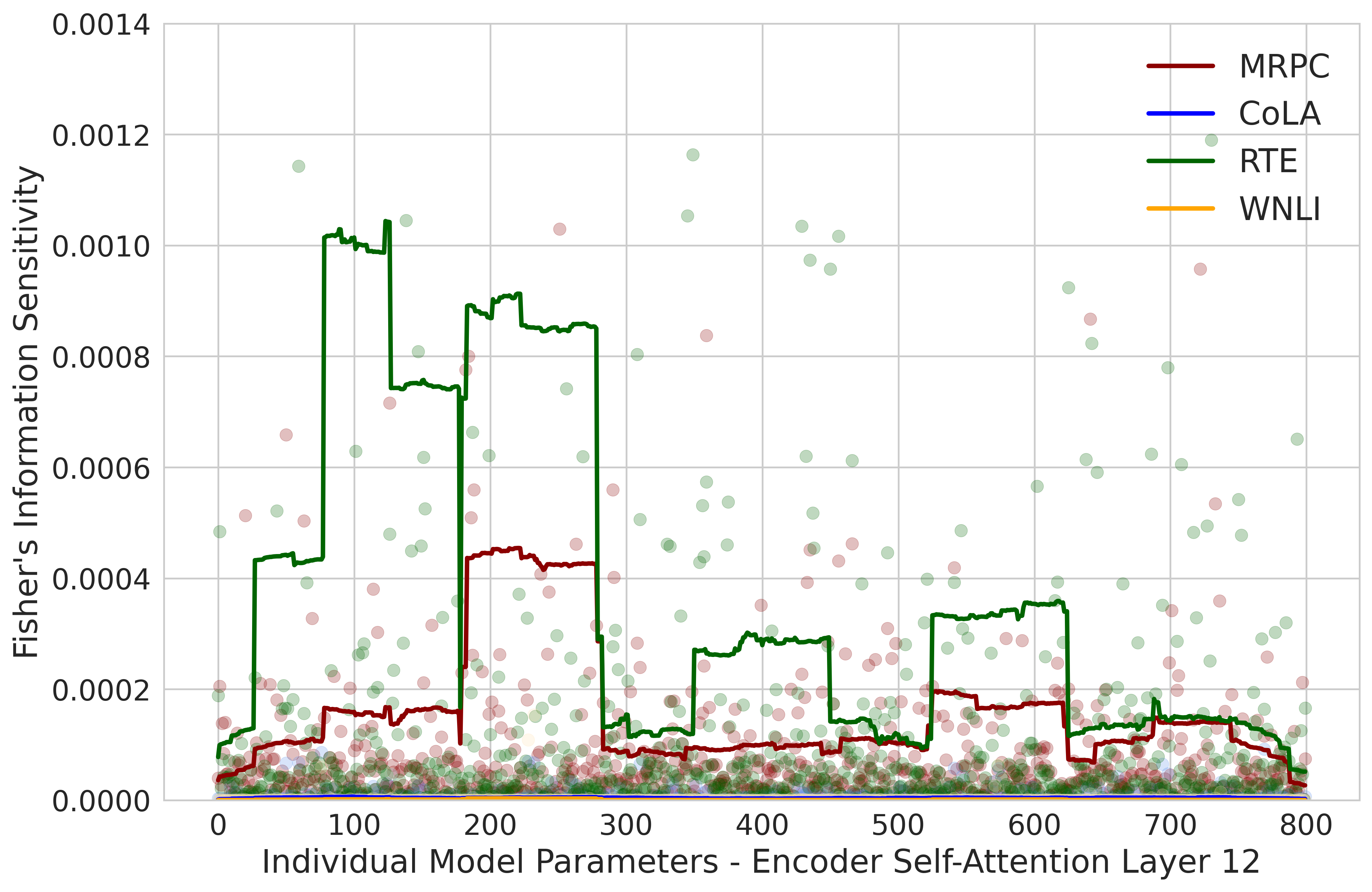}
    \caption{\textit{NLU tasks MRPC, RTE, and CoLA, suffer the most as a consequence of non-linguistic skill-injection,  because these tasks deem the same subset of model parameters to be vital as  the non-linguistic (arithmetic) task:} Given Fisher parameter sensitivities $I(\theta)$ of the first (left) and last (right) self-attention encoder layers for four different models based on continued training of the base BERT model on four datasets: $I_{arith}(\theta)$ on an arithmetic reasoning and $I_{CoLA}(\theta)$, $I_{MRPC}(\theta)$, $I_{RTE}(\theta)$ on GLUE tasks CoLA, MRPC, and RTE respectively, this plot takes the $n=800$ most crucial parameters based on $I_{arith}(\theta)$ and showcases how sensitive those \textit{same} parameters are to the GLUE tasks based on $I_{CoLA}(\theta)$, $I_{MRPC}(\theta)$, and $I_{RTE}(\theta)$. These plots correlate to the task-specific performance degradation showcased in Table \ref{table:1}.}%
    \label{fig:fisherlayers}
\end{figure*}

\vspace{0.2cm}
\noindent\textbf{Substantiating forgetting on the basis of parameter sharing:} To establish that observed performance degradation can indeed be accredited to catastrophic forgetting, we take an information theoretic lens to pry into parameter-sharing tendencies across tasks with the aid of Fisher information \citep{fisher}. For a single sample $y$ drawn from a distribution with probability desnity $f(y;\theta)$, the Fisher information index $I(\theta)$ (\ref{fisher}) quantifies the sensitivity of the parameter $\theta$ to the data instance $y$. Thus, given a task-specific training corpus $(X, Y) \in D_{task}$, we can estimate the sensitivity of each model parameter $\theta_{i} \in \theta$ for the given task.
\begin{align}
\footnotesize
\label{fisher}
    I(\theta_{i}) &= E_{y \in Y}(\frac{d \log_{} f(y;\theta_{i})}{d\theta_{i}})^{2} \\
    &= -E_{y \in Y}(\frac{d^{2} \log_{} f(y;\theta_{i})}{d\theta_{i}^{2}}) 
\end{align}
Using this formulation, we compute the Fisher parameter sensitivities $I(\theta)$ for four different models based on continued training of the base BERT model on four datasets:
\begin{itemize}
    \item $I_{arith}(\theta)$: for BERT trained on an arithmetic reasoning dataset \cite{geva:2020}
    \item $I_{CoLA}(\theta)$, $I_{MRPC}(\theta)$, $I_{RTE}(\theta)$: for BERT trained on three GLUE \citep{glue} tasks CoLA, MRPC, and RTE respectively
\end{itemize}
To ground our hypothesis of task-specific forgetting as a consequence of parameter-sharing, first, we select $n=800$ parameters deemed most sensitive for arithmetic reasoning from $I_{arith}(\theta)$, and compare how important those \textit{same} parameters are for the three GLUE tasks based on their respective Fisher scores $I_{CoLA}(\theta)$, $I_{MRPC}(\theta)$, $I_{RTE}(\theta)$ (see Appendix \textcolor{blue}{\textsection A.1.1} for details on Fisher score computations). As seen in Figure \ref{fig:fisherlayers}, for the first and last self-attention encoder layers, the sensitivities of the parameters across tasks correlate well with the findings of Table \ref{table:1} - the NLU task that suffers the least performance degradation (WNLI) also has the least sensitivity to these (shared) parameters across the encoder layers, while the NLU tasks that do suffer from performance degradation (MRPC, CoLA, RTE) have varying ranges of shared sensitivities across the encoder self-attention layers. These findings hold consistent across all 12 encoder layers of the BERT model (see Appendix \textcolor{blue}{\textsection A.1.2}). 
\vspace{0.25cm}

\noindent \textbf{Our contributions:} In line with the above observations, we offer the following contributions in the form of our proposed model, \textit{\model}, for non-linguistic skill injection in LLMs:
\begin{itemize}
    \item Novel multi-task skill-injection loss that infuses a sense of numeral structure in the learned representations, leading to better generalization performance than the state-of-the-art, all with a significantly lower fraction ($\frac{n}{4}$) of training data. 
    \item Weight consolidation schemes for LLMs for better linguistic retention with \textit{0} additional linguistic samples compared to \textit{1 million} synthetic textual training samples used by the state-of-the-art.
    \item Through exhaustive qualitative and quantitative evaluations, we demonstrate the improved generalization performance of {\model} over the state-of-the-art. Our experiments also highlight the need for disclosing linguistic performance for models trained on highly-niche non-linguistic tasks.
\end{itemize}

\section{Designing \model}
\begin{figure*}[t]
    \centering
    \includegraphics[scale=0.28]{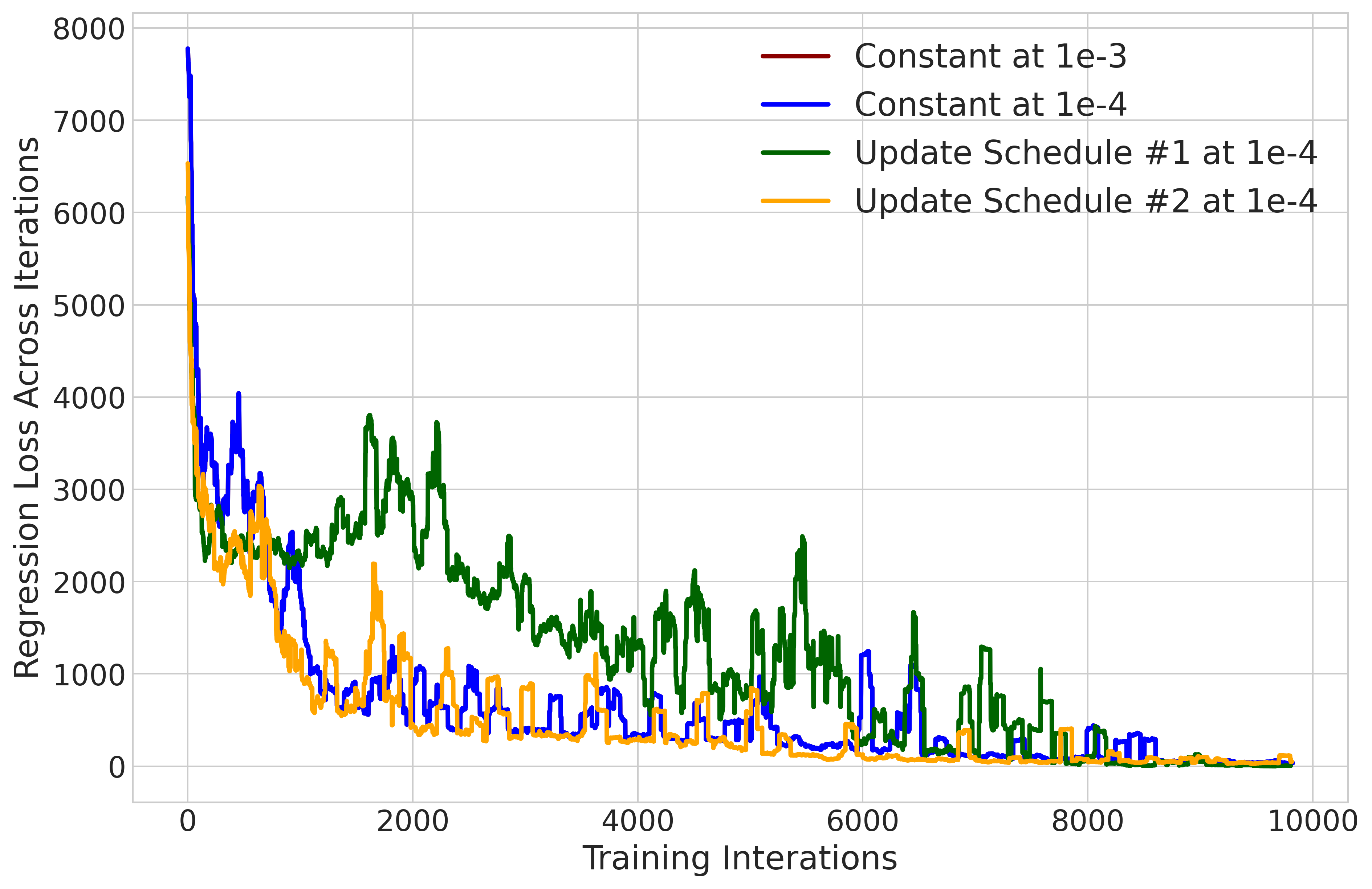}
    \includegraphics[scale=0.28]{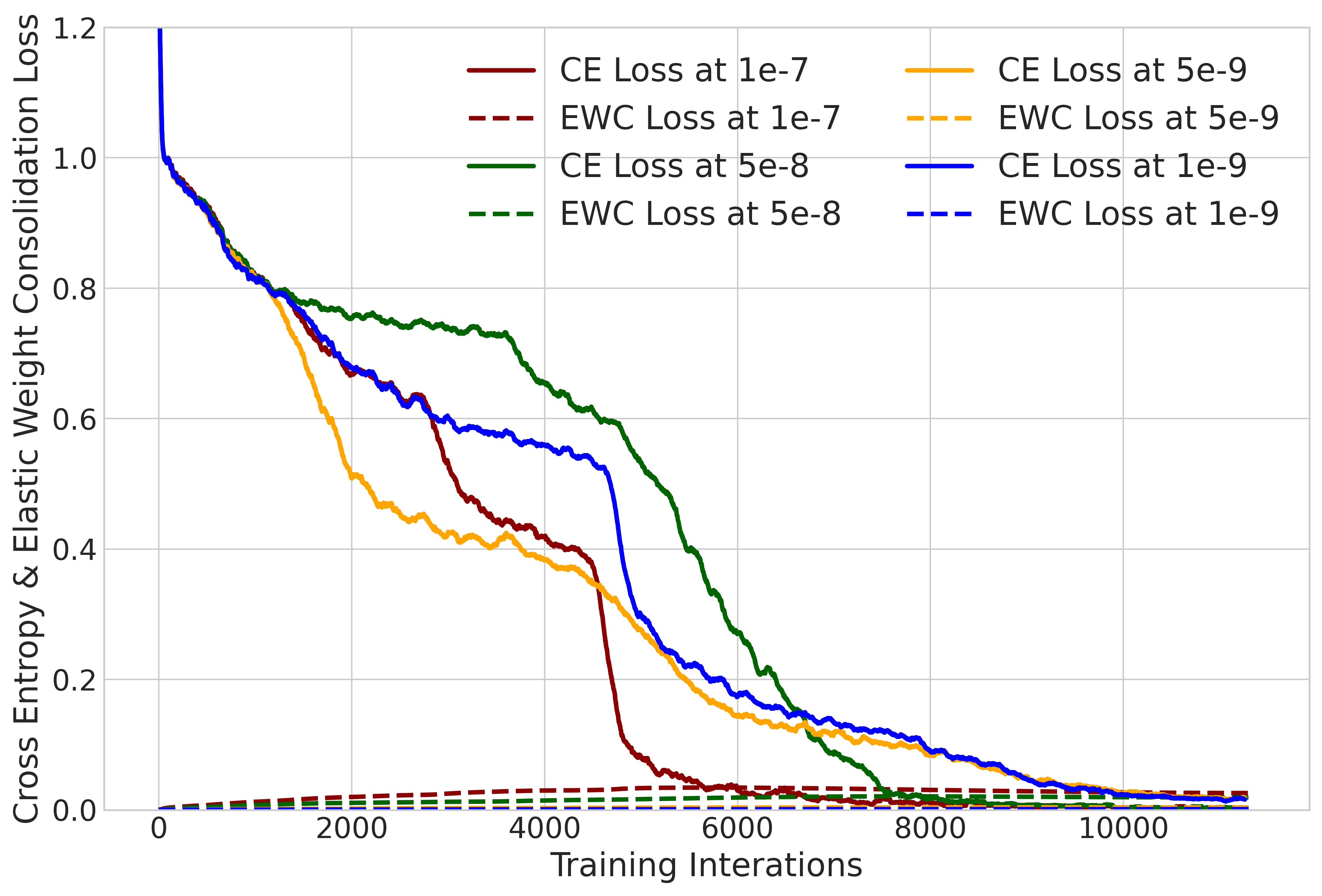}
    \caption{Empirical setting of hyperparameters: For regression loss $\mathcal{L}_{REG}$ (left), the loss convergence is evaluated at four configurations of $\lambda_{1}$ on the validation set with both constant $\{1e^{-3},1e^{-4}\}$ and dynamic (update scheduling) initializations. For EWC loss $\mathcal{L}_{EWC}$ (right), the interplay between $\mathcal{L}_{EWC}$ and CE loss $\mathcal{L}_{CE}$ (color-matched) is evaluated through a parameter sweep of $\lambda_{2}$ within $\{1e^{-6},1e^{-10}\}$ with the best performing configurations plotted.}%
    \label{fig:losses}
\end{figure*}

\subsection{Non-Linguistic Learning}

Based on probabilistic modeling, language models are trained to output the next sequential token $y_{t}$ at timestep $t$ based on the $n$ tokens already predicted by the model, formulated as $P(y_{t} | y_{t-1},...,y_{t-n}) = P(y_{t} | y_{<t})$. This probability distribution $P$ is often optimized through measures of uncertainty such as cross-entropy or KL-divergence. The application of these same loss functions used for learning linguistic token distributions may not necessarily translate to the learning of non-linguistic entities.

Unlike linguistic tokens, the magnitude of a numeral is especially tied to its meaning \citep{dehaene:1998}. This magnitude can either be modeled as a continuous linear representation \cite{dehaene:1990} or a log-compressive representation \citep{dehaene:2003}. Thus, to inject this numeric-scale representation into a language model, we take a simplistic approach of augmenting the learning of tokens through cross-entropy \textit{$\mathcal{L}_{CE}$} with a regression loss \textit{$\mathcal{L}_{REG}$}\footnote{As in the initial phases of model training, incorrect predictions of target numerals can lead to exceedingly large values of $\mathcal{L}_{REG}$, thus our choice of seed values for $\lambda_{1}$ were set to $\{1e^{-3},1e^{-4}\}$ as not to exceed the range of $\mathcal{L}_{CE}$ $\{0,1\}$ by values greater than an order of magnitude.}. This regression loss is incorporated into the quantitative reasoning loss function \textit{$\mathcal{L}_{Q}$} as represented in (\ref{regression:1}, \ref{regression:2}).
\begin{align}
\label{regression:1}
    \mathcal{L_{Q}}(\theta) &= \mathcal{L}_{CE} + \lambda_{1} \, . \, \mathcal{L}_{REG} \\ 
\label{ce:1}
    \mathcal{L}_{CE} &= -log(P(y_{t} | y_{<t})) \\
\label{regression:2}
    \mathcal{L}_{REG} &= \sqrt{\sum_{i=1}^{n} (y^2 - \hat{y}^2)}
\end{align}

\noindent Figure \ref{fig:losses} (left) depicts the convergence of $\mathcal{L}_{REG}$ for different configurations of $\lambda_{1}$. Please see Appendix \textcolor{blue}{\textsection A.2.1} for further details on the update schedules for hyperparameter tuning.

\subsection{Linguistic Retention}

Among prominent strategies for multitask learning, a system-level consolidation scheme consists of stitching-together amalgamated datasets constituting multiple-shared tasks \cite{kumaran:2016}. However, due to the limitless range of possible downstream tasks that LLMs are often employed for, the paradigm consists of building large models that hold linguistic prowess and are intended to be fine-tuned on a single downstream task \cite{bert, gpt2}, thus suited for a continual learning paradigm. As depicted in Figure \ref{fig:examples}, the high degree of parameterization of these models leads to the belief that there is a solution for \{task $B$, $\theta_{B}$\}, a non-linguistic skill, that is proximal to the linguistic solution space for the model \{task $A$, $\theta_{A}$\} \citep{sharma:2022_2}. To enable this continual learning, we adapt the elastic weight consolidation (EWC) regularization to LLMs - elastic as it functions as a spring, anchoring the solution space closer to $\theta_{A}$ \citep{kirkpatrick:2017}. Thus, EWC penalizes changes to specific network weights deemed vital for linguistics while injecting non-linguistic skills into the model.

In line with our task-specific parametric observations from Introduction \textcolor{blue}{\textsection 1.2}, we compute $ F = I_{BERT}(\theta)$, the Fisher information index for the base BERT model based on a portion of its original pre-training corpus - WikiText \citep{wikitext}, thus approximating its posterior distribution. Let us assume that $\theta^{*}_{ling}$ represents the set of parameters of a converged base-BERT model pre-trained for \textit{linguistics}. We now introduce the quadratic penalty $\mathcal{L}_{EWC}$ (\ref{loss:1}, \ref{loss:2}) that penalizes changes to any model parameter $i$ crucial to the core linguistic functionality of the pre-trained model. 
\begin{align}
\label{loss:1}
    \mathcal{L}(\theta) &= \mathcal{L}_{Q}(\theta) + \lambda_{2} \, . \, \mathcal{L}_{EWC} \\
\label{loss:2}
    \mathcal{L}_{EWC} &= \sum_{i} \frac{1}{2} \ F_{i}(\theta_{i} - \theta_{ling,i}^{*})^{2}
\end{align}
In this loss formulation, the hyperparameter $\lambda_2$ is crucial as it dictates both model convergence and balances the learning of quantitative reasoning skills $\theta_{Q}$ with linguistic prowess $\theta_{ling}$. To evaluate the sensitivity of model convergence with respect to $\lambda_{2}$, we perform a hyperparameter sweep between $\{1e^{-6},1e^{-10}\}$ - Figure \ref{fig:losses} (right) showcases the interplay between $\mathcal{L}_{CE}$ and $\mathcal{L}_{EWC}$ (color-matched) for the best performing values of $\lambda_{2}$ on the validation set. The first sign of model convergence is observed at $\lambda_{2} = 1e^{-7}$, and although slight improvements to model convergence are noted for even smaller values of $\lambda_{2}$, the smallest value that allows for convergence, theoretically, allows for balanced learning of $\theta_{Q}$ with $\theta_{ling}$.

\section{Experiment Setup and Results}
\subsection{Tasks and Datasets}
The goal of {\model} is to empower LLMs with non-linguistic skills in a manner that avoids catastrophic forgetting of linguistic skills without the aid of additional synthetic linguistic training. Thus, we have two categories of tasks that \model, along with the baselines, should be evaluated on:
\subsubsection{Quantitative Reasoning}
To hold fair comparisons to GenBERT \cite{geva:2020}, we both train and evaluate all models with the arithmetic reasoning portion of their dataset. The data instances take the form of the sample arithmetic task demonstrated in Figure \ref{fig:examples}. The corpus consists of $N_{train} = 165,000$ training samples and $N_{val} = 1666$ validation samples, where the numerals are in the range $\{1,20^{3}\}$ with numeral ranges stratified between the training and validation sets. For our models, we randomly sample $\frac{N_{train}}{4}$ instances for training.
\vspace{0.2cm}
\par\noindent
\textbf{OOD Performance}: The out-of-domain (OOD) performance of all models are evaluated on data instances generated in the same manner but for numeral ranges $\{20^{3},10^{6}\}$ that are unseen for all models evaluated.
\subsubsection{Natural Language Understanding}
Following standard protocols, we employ all 9 tasks in the GLUE benchmark \cite{glue} as metrics for linguistic prowess of a model. The tasks, as per the benchmark, are categorized into three groups:
\begin{itemize}
    \item Single Sentence Tasks: CoLA (the Corpus of Linguistic Acceptability) \cite{cola} for grammatical fidelity (Matthews correlation), SST-2 (the Stanford Sentiment Treebank) \cite{sst2} for sentiment prediction
    \item Similarity and Paraphrase Tasks: MRPC (the Microsoft Research Paraphrase Corpus) \cite{mrpc}, QQP (the Quora Question Pairs), and STS-B (the Semantic Textual Similarity Benchmark) \cite{stsb} for semantic equivalence
    \item Inference Tasks: MNLI (the Multi-Genre Natural Language Inference Corpus) \cite{mnli} and RTE (Recognizing Textual Entailment) for textual entailment, QNLI (the Stanford Question Answering Dataset) \cite{squad} for Q\&A, and WNLI (the Winograd Schema Challenge) \cite{wnli} for pronoun referent selection.
\end{itemize}

\par\noindent
\textbf{Evaluation Metrics}: Besides CoLA (evaluated using the Matthews correlation coefficient) and STS-B (evaluated using a combination of the Spearman's and Pearson's correlation coefficients), all result shown represent accuracy for the respective GLUE task.

\subsection{Baselines}

\begin{table*}[t]
\small
\centering
\begin{tabular}{@{}lccccc@{}}
\toprule
                          &                       & \multicolumn{4}{c}{Model Accuracy}                                         \\ \midrule
Model                     & Training Samples      & Validation Set {[}$0$,$20^3${]} & OOD {[}$20^{3}$,$10^{4}${]} & OOD {[}$10^4$,$10^5${]} & OOD {[}$10^5$,$10^6${]} \\ \midrule
GenBERT                   & 165,000 (n)           & \textbf{100\%} & 1.32\%            & 0.06\%            & 0.0\%             \\
BERT                      & \textbf{41,250 (n/4)} & 96.63\%        & 7.20\%            & \textbf{0.12\%}   & 0.0\%             \\
{\model}  (w/o $\mathcal{L}_{EWC}$)       & \textbf{41,250 (n/4)} & 95.67\%        & 9.66\%            & \textbf{0.12\%}   & 0.0\%             \\
\model  & \textbf{41,250 (n/4)} & 98.01\%        & \textbf{19.44\%}  & \textbf{0.12\%}   & 0.0\%             \\ \bottomrule
\end{tabular}
\caption{Comparative analysis of quantitative reasoning performance between the baselines and {\model} (that uses $\frac{1}{4}$ of the training data) for both the in-domain validation set $\{1,20e^{3}\}$ and out-of-domain (OOD) sets for numeral ranges $\{20e^{3},10e^{4}\}$, $\{10e^{4},10e^{5}\}$, and $\{10e^{5},10e^{6}\}$. Using a fraction ($\frac{n}{4}$) of the original training data, {\model} not only closely matches the in-domain performance of GenBERT, it significantly improves the generalization performance to OOD numerals ranges as well.}
\label{table:2}
\end{table*}
\begin{table*}[t]
\small
\centering
\begin{tabular}{lcccccc}
\hline
Model        & Training Samples & CoLA                & STS-B               & MNLI                 & MNLI$_{MM}$            & MRPC                 \\ \hline
BERT         & -                & \textit{0.59}       & \textit{0.89}       & \textit{83.85}       & \textit{84.05}       & \textit{86.76}       \\
BERT$_{Arith}$ & \textbf{0}       & 0.08                & 0.80                & 32.73                & 32.95                & 70.34                \\
GenBERT      & 1 Million        & $0.54_{0.001}$          & $0.88_{0.001}$          & $83.00_{0.576}$          & $83.40_{1.107}$          & $85.04_{0.693}$          \\
\model    & \textbf{0}       & \textbf{0.58}$_{0.041}$ & \textbf{0.89}$_{0.003}$ & \textbf{84.07}$_{0.158}$ & \textbf{84.66}$_{1.123}$ & \textbf{86.88}$_{1.123}$ \\ 
\end{tabular}
\begin{tabular}{@{}lcccccc@{}}
\toprule
Model        & Training Samples & QNLI                 & QQP                  & RTE                  & SST-2                & WNLI                 \\ \midrule
BERT         & -                & \textit{90.55}       & \textit{90.61}       & \textit{65.34}       & \textit{91.62}       & \textit{56.33}       \\
BERT$_{Arith}$ & \textbf{0}       & 50.53                & 70.49                & 47.29                & 88.07                & 56.33                \\
GenBERT      & 1 Million        & 90.83$_{0.012}$          & 90.78$_{0.316}$          & \textbf{67.86}$_{2.042}$ & 91.51$_{0.648}$          & 55.63$_{0.995}$          \\
\model    & \textbf{0}       & \textbf{91.54}$_{0.207}$ & \textbf{90.96}$_{0.043}$ & 65.70$_{1.531}$          & \textbf{92.37}$_{0.081}$ & \textbf{56.18}$_{0.216}$ \\ \bottomrule
\end{tabular}
\caption{Comparative analysis of linguistic performance between the baselines and {\model} (that uses 0 additional linguistic training data) for the set of 9 GLUE benchmarks, each addressing a certain aspect of natural language understanding (see \textcolor{blue}{\textsection 3.1.2}). To further authenticate the performance increase of {\model} over GenBERT, the results are presented as $\mu_{\sigma}$ (mean and standard deviation) across two runs of training-validation with different seeds for model initialization.}
\label{table:3}
\end{table*}

For assessment of both quantitative reasoning skills and linguistic prowess through natural language understanding, the following three models are used as the baselines for this experimentation. For the training specifics, please see Appendix \textcolor{blue}{\textsection A.2.2}.
\begin{itemize}
    \item \textit{BERT}: In this evaluation, this base pre-trained model establishes the standard for natural language understanding that all BERT-derivatives designed for non-linguistic skills should strive to achieve. Thus, its performance on the set of GLUE tasks are \textit{italicized} in Table \ref{table:3}.
    \item \textit{BERT$_{Arith}$}: This is the model generated from the continued training of the pre-trained BERT model on the quantitative reasoning dataset using the standard cross-entropy $\mathcal{L}_{CE}$ loss. This model showcases the current paradigm of skill-injection where the architecture of a model is left unchanged and the training parameters are often adapted to meet performance requirements in the target task.
    \item \textit{GenBERT} \cite{geva:2020}: This BERT-based model is trained for numerical reasoning with a multitask setup wherein a conjunction 1 million synthetic numerical reasoning samples (165,000 of which are strict arithmetic) is used for numeric skill injection while an additional 1 million synthetic textual samples are used to avoid catastrophic forgetting of linguistics as a consequence of the non-linguistic skill injection. Please note, that for this experimentation, the \textit{pre-trained GenBERT model has been used as-is, thus ensuring no performance degradation as a consequence of in-house replication}.
\end{itemize}

\subsection{Quantitative Results}
\subsubsection{Numerical Reasoning}

From Table \ref{table:2}, we observe that using only $\frac{1}{4}$th of the training dataset, {\model} closely resembles the performance of GenBERT while significantly improving the performance on out-of-domain numeral ranges. This leads to two deductions:
\begin{itemize}
    \item It is known in the literature that LLMs often struggle to extrapolate numeral ranges that are absent from the training corpus (OOD) \citep{wallace:2019,razeghi:2022}. The significant improvement in quantitative reasoning in OOD numerals from {\model} (w/o $\mathcal{L}_{EWC}$) (row 3) establishes the vital role that skill-specific regression loss $\mathcal{L}_{REG}$ plays in not just learning the correct tokens to predict in response to a quantitative prompt, but capturing the magnitude of each numeral tokens in their representations.
    \item The significant jump in OOD improvement in addition to the increased in-domain performance from {\model} (row 4) suggests that $\mathcal{L}_{EWC}$ not only minimizes the loss of linguistic prowess, but also acts as a universal regularizer that prevents the model from over-fitting on the target task.
\end{itemize}

\begin{figure*}[t]
    \centering
    \includegraphics[scale=0.5]{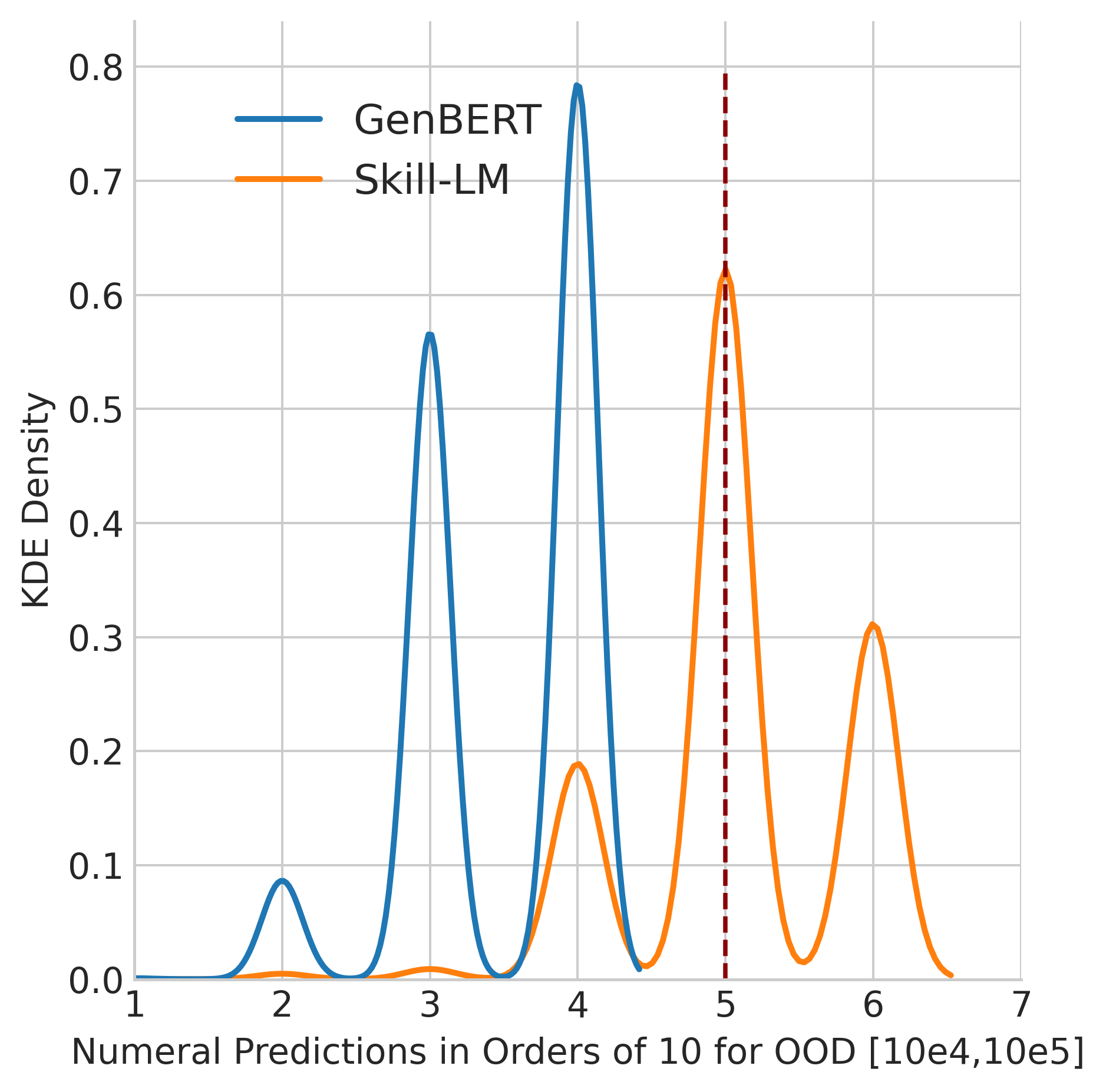}
    \includegraphics[scale=0.5]{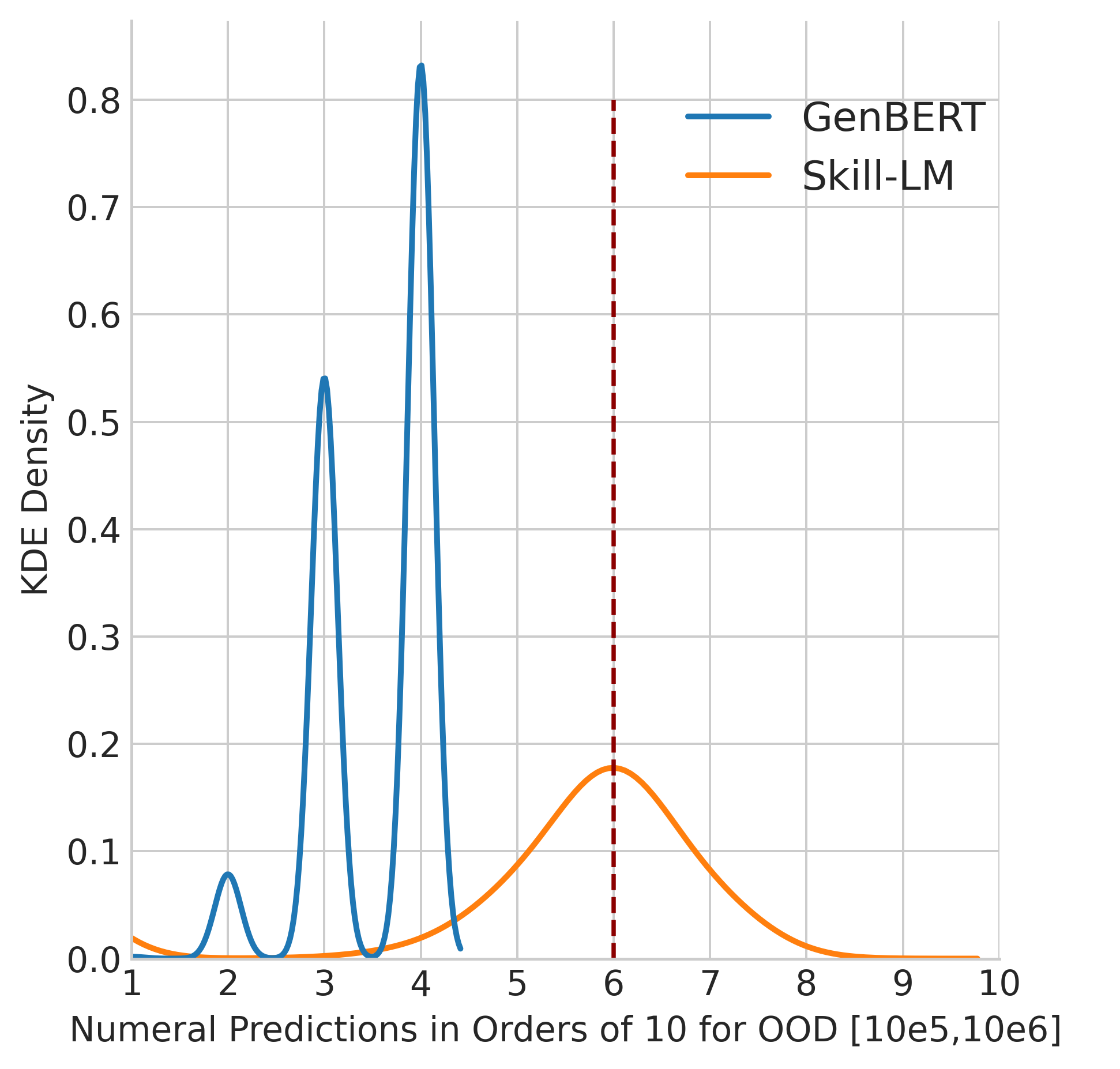}
    \caption{\textit{{\model} consistently predicts the correct order of magnitude for the numerals:} From Table \ref{table:2}, {\model} significantly improves predictive performance for OOD range $[20^3,10^4]$, however, all models see a drastic drop in performance for the OOD ranges $[10^4,10^5]$ and $[10^5,10^6]$. Although these models are unable to predict the \textit{exact} numerals, the KDE plots above showcase how \textit{close} these models are to predicting the correct order of magnitude for the numerals - $[10^4,10^5]$ on the left and $[10^5,10^6]$ on the right. {\model} consistently predicts the correct order of magnitude for the numerals as marked by the vertical dashed red line. This is evident from the fact that the largest mode of {\model} coincides with the correct order of magnitude (red line).}%
    \label{fig:magnitudes}
\end{figure*}

\subsubsection{Natural Language Understanding}
Recall that our goal with {\model} is to prevent the loss of linguistic prowess as a consequence of non-linguistic skill injection. The premise therein is that BERT-derivatives, empowered with non-linguistic skills, should at least strive to have linguistic performances of the base model. Thus, the performance of the base BERT model is \textit{italicized} in Table 3.

In \textcolor{blue}{\textsection 1.2}, we established the degradation of linguistic performance in LLMs as a consequence of non-linguistic skill injection. Thus, the goal of weight consolidation $\mathcal{L}_{EWC}$ was to revitalize the linguistic performance of the model back to baseline. However, from Table \ref{table:3}, we observe that employing $\mathcal{L}_{EWC}$ that uses 0 additional training data outperforms GenBERT that uses 1 Million additional linguistic training data on 8/9 of the standardized GLUE benchmarks. To further authenticate these findings, the results are presented as $\mu_{\sigma}$ (mean and standard deviation) across two runs of training-validation with different seeds for model initialization. Thus {\model} showcases improved performance coupled with significant \textit{savings in GPU compute costs} compared to previous related efforts that
train on an additional 1 Million linguistic training samples~\cite{geva:2020}.

\vspace{-0.1cm}
\subsection{Qualitative Results}

In \textcolor{blue}{\textsection 2.1}, we theorized that regression loss, in the context of numerical skill injection, would inject a sense of numeric scale and magnitude estimation \citep{dehaene:1998} to the general learning of numerical representations. From Table \ref{table:2} we quantified the gains from this skill-specific loss in OOD generalization of numerals, however, in this section we further investigate whether the extrapolation to OOD numerals is indeed due to this learnt sense of numeric scale. 

\begin{figure}[h]
    \centering
    \caption{For the OOD range $[20^3,10^4]$ immediate to the training numeral range $[0,20^3]$, this figure showcases, qualitatively, the predictive behaviors of {\model} vs GenBERT. Although GenBERT is able to capture the nuances in difference of the operands, it fails to extrapolate to the scale of the operands.}%
    \includegraphics[scale=0.35]{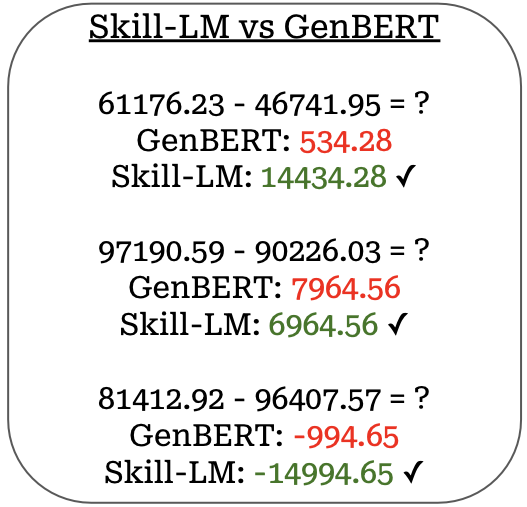}
    \label{fig:samples}
\end{figure}

\noindent\textbf{OOD numerals closer to the training range:} In Table \ref{table:2}, {\model} boosts the predictive performance for OOD numerals in the range $[20^3, 10^4]$ from 1.32\% to 19.44\% - but where does the baseline fail? In Figure \ref{fig:samples}, as common-case failure scenarios, we showcase 3 sample responses from {\model} vs baseline to prompts from the OOD range $[20^3, 10^4]$: while the baseline does capture the nuances in difference of the operands (the numerals closer the decimal are correct), it severely fails to extrapolate to the scale of the operands.

\noindent\textbf{OOD numerals further from the training range:} In Table \ref{table:2}, the evaluation metric used is accuracy, thus evaluating the capabilities of these models to output the \textit{exact} token in response to the quantitative reasoning prompts. For larger OOD ranges $[10^4,10^5]$ and $[10^5,10^6]$, all models struggle to predict the exact output - \textit{but how close do they get?}
Figure \ref{fig:magnitudes} showcase the distribution of the predicted output based on their powers of 10s - for the OOD range $[10^4,10^5]$, the outputs should mostly center around $10^5$ (left figure) while for the range $[10^5,10^6]$ they should center around $10^6$ (right figure). Although unable to predict the exact tokens, {\model} tends to predict tokens closer in magnitude to the ground truth consistently compared to our baseline.

\section{Conclusions}
Our study shows that LLMs are capable of demonstrating quantiative reasoning without sacrificing the broad palette of linguistic skills that they are traditionally evaluated against. This multi-task framework, together with the weight consolidation strategy, highlights that this framework can be systematized beyond the studies described here. As a result, non-linguistic tasks and linguistic tasks need not be seen as being at odds for LLMs and we can begin thinking about richer integrations of qualitative and quantitative reasoning. Our experimental results also highlight that the improvements showcased here do not require exorbitant training data and in fact require just a fraction of what previous studies have leveraged.

Our future work will be organized in three directions. First, we intend to study at a more fine-grained level the dovetailing of different arithmetic reasoning tasks vis-a-vis linguistic counterparts, and any synergies that can be exploited while learning. Second, there are situations where linguistics can help numerical reasoning (math word problems, data-to-text generation) and multi-task formulations that capture the underlying semantics can be developed. Finally, there are other forms of non-linguistic reasoning (diagrammatic reasoning) that can potentially be studied using the multi-task framework that we have described here.

\section*{Limitations}
In our study, we address the issue of linguistic forgetting via the injection of the strict non-linguistic skill of quantitative reasoning. Although quantitative reasoning with LLMs is an active research area, as discussed above, further fine-grained studies are required to extrapolate this behavior to tasks that leverage synergies between aspects of both linguistics and non-linguistics - such as math word problems or data-to-text generation. Further, investigations into the linguistic forgetting tendencies of different languages would lend an insight into the role of linguistic morphology in this behavior. The restrictions from our in-house GPU resources does not allow scaling this study to more recent models that exceed 100 Billion parameters, although, due to the sharing of similar architectures, we forecast our findings to hold despite of model scaling.

\section*{Ethics Statement}
Although the ethical waters of the development and deployment of LLMs are difficult to nagivate, we can ascertain that our study does not bring forth further complications. The datasets we use in this study are established benchmark datasets from publicly accessible websites and do not contain any personally identifiable information. Our analyses does not constitute human subjects and thus is not within the purview of the IRB. Further, in the landscape of increasing emission costs from large-scale computation, our study offers avenues for severely restricting the size of the training data - both linguistic and non-linguistic.

\bibliography{custom}
\bibliographystyle{acl_natbib}

\appendix

\section{Appendix}
\label{sec:appendix}
\subsection{Substantial Forgetting on the Basis of Parameter Sharing}

\subsubsection{Fisher Information Computation}
The Fisher information score, as depicted in (\ref{fisher}) is the expected value of the square of the gradient for a sample $y \in Y$. Thus, to compute the Fisher sensitivity of a model $\theta$ to a task $A$, we compute the sum of the squared gradients averaged by the number of parameters in $\theta$. In our case, where $\theta$ is a pretrained transformer-based LLM, the model cross-entropy loss $dlog f(y;\theta)$ (\ref{ce:1}) for each sample $y$ is computed, through which the gradient $\frac{dlog f(y;\theta)}{d\theta}$ can then be computed. The sum of these squared gradients gives us the Fisher information score for each parameter $\theta_{i}$ in a model $\theta$ with respect to a task $A$.

\subsubsection{Parameter Sensitivities for the Self-Attention Encoder Layers}
\begin{figure*}[h]
    \centering
    \includegraphics[scale=0.22]{images/fisher/1.png}
    \includegraphics[scale=0.22]{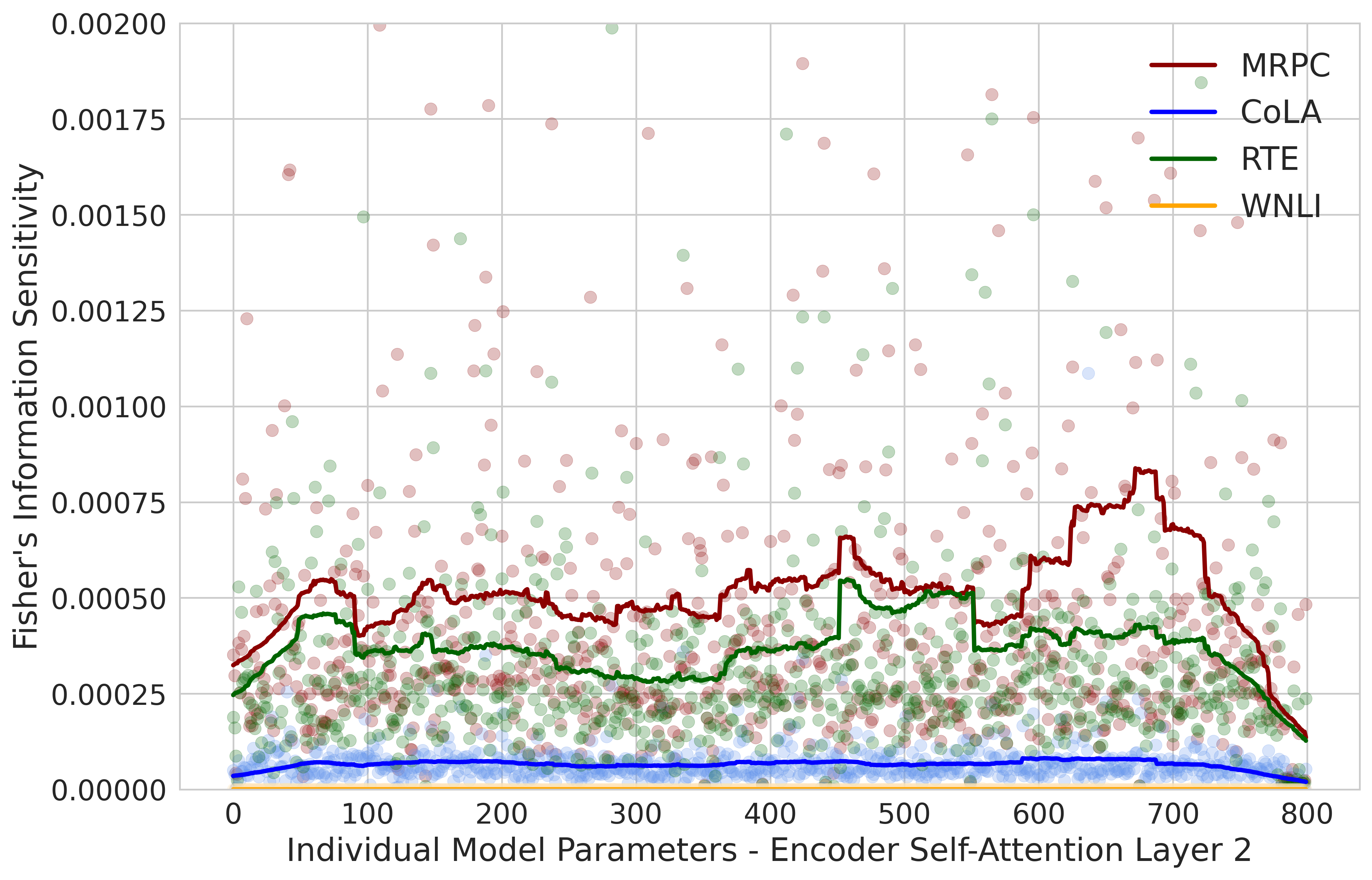}
    \includegraphics[scale=0.22]{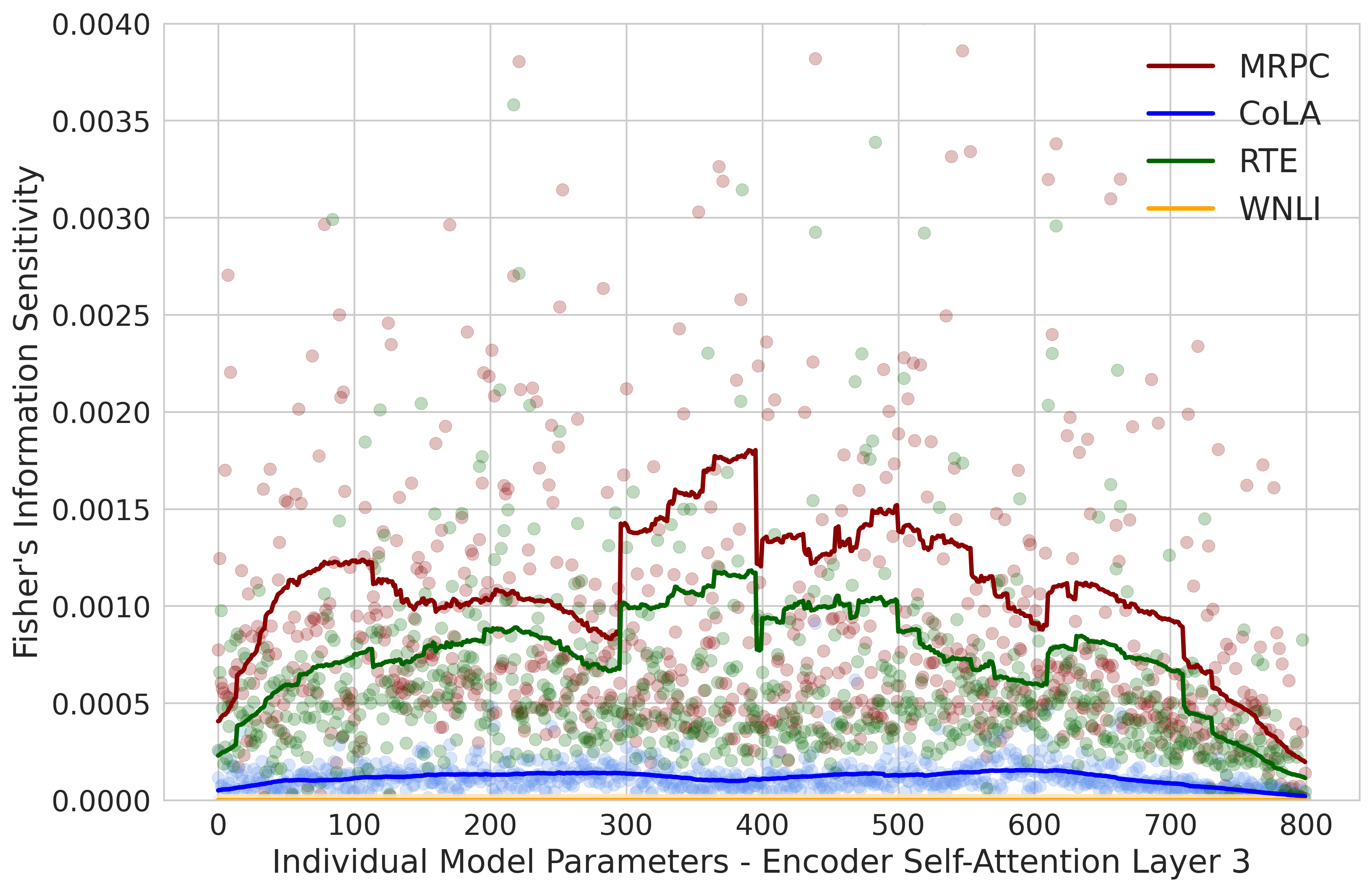}
    \includegraphics[scale=0.22]{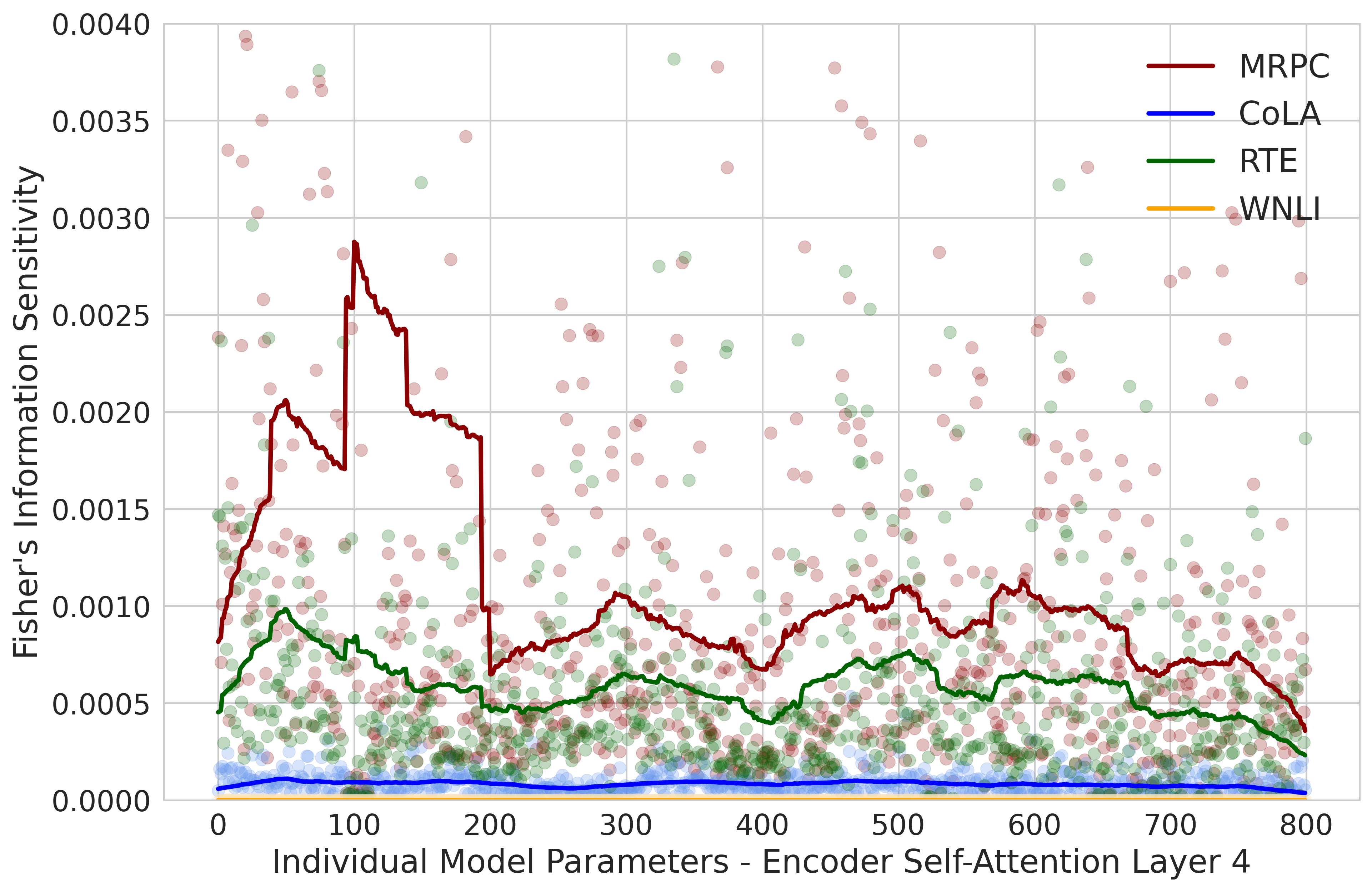}
    \includegraphics[scale=0.22]{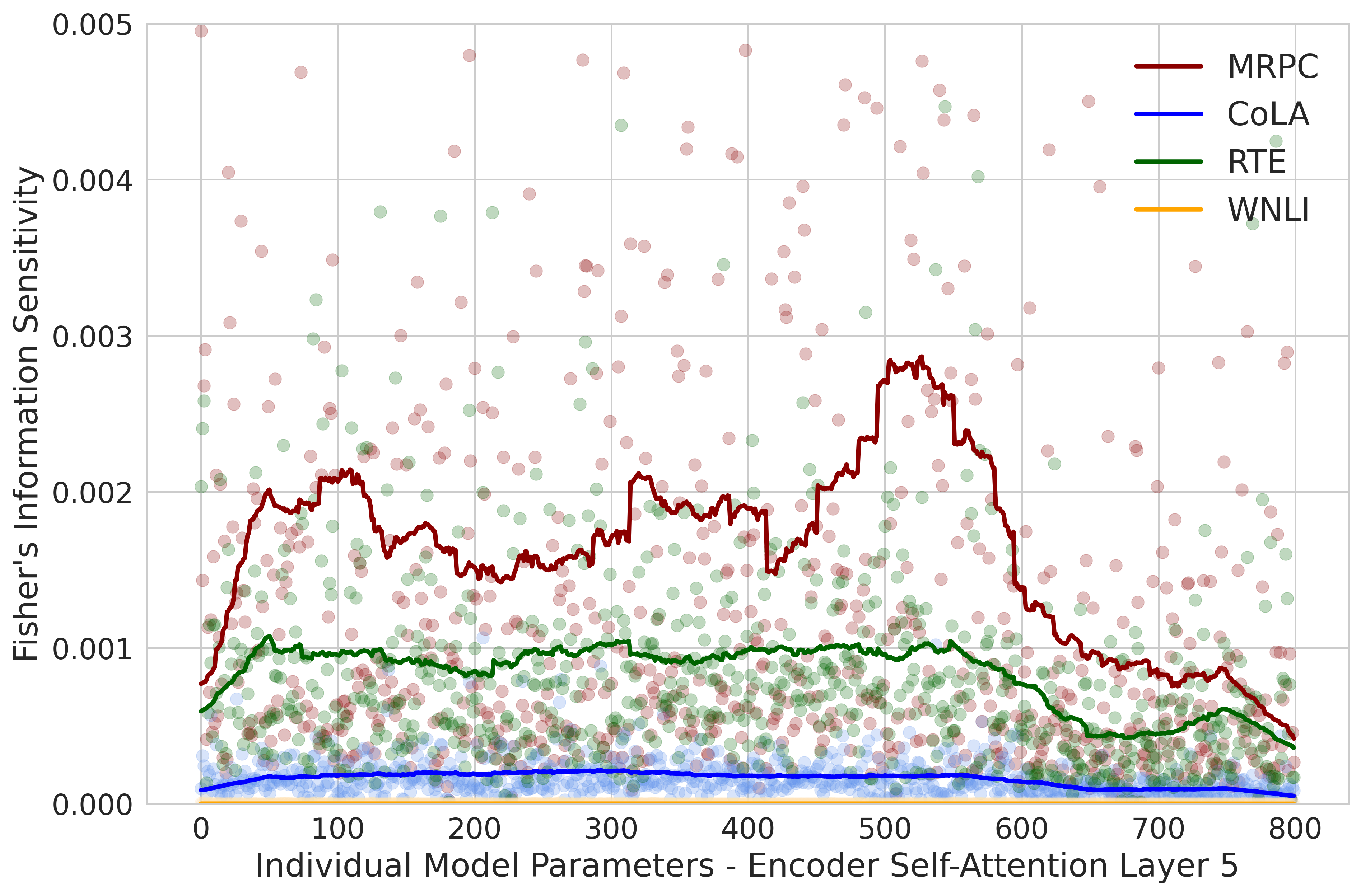}
    \includegraphics[scale=0.22]{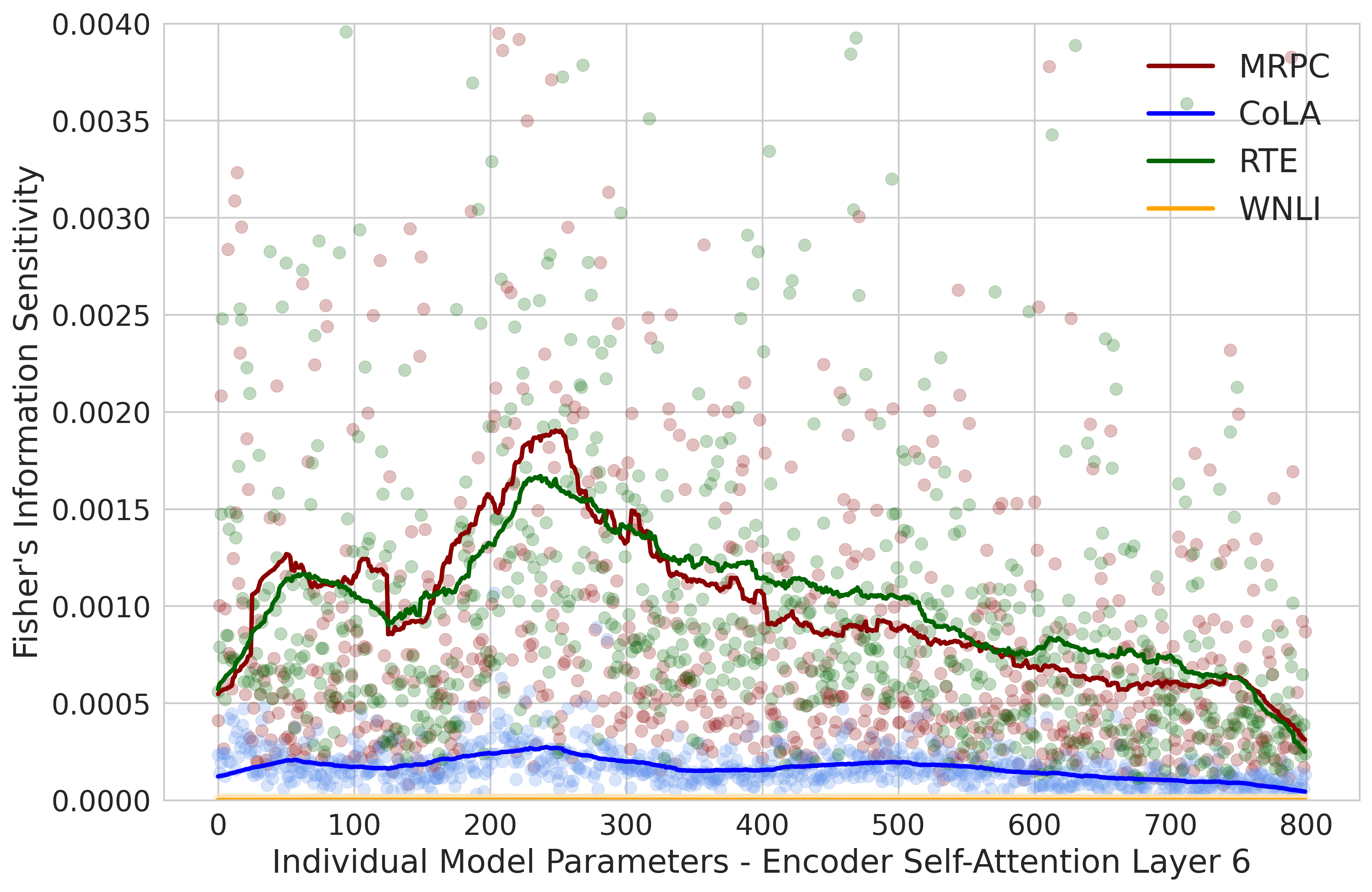}
    \includegraphics[scale=0.22]{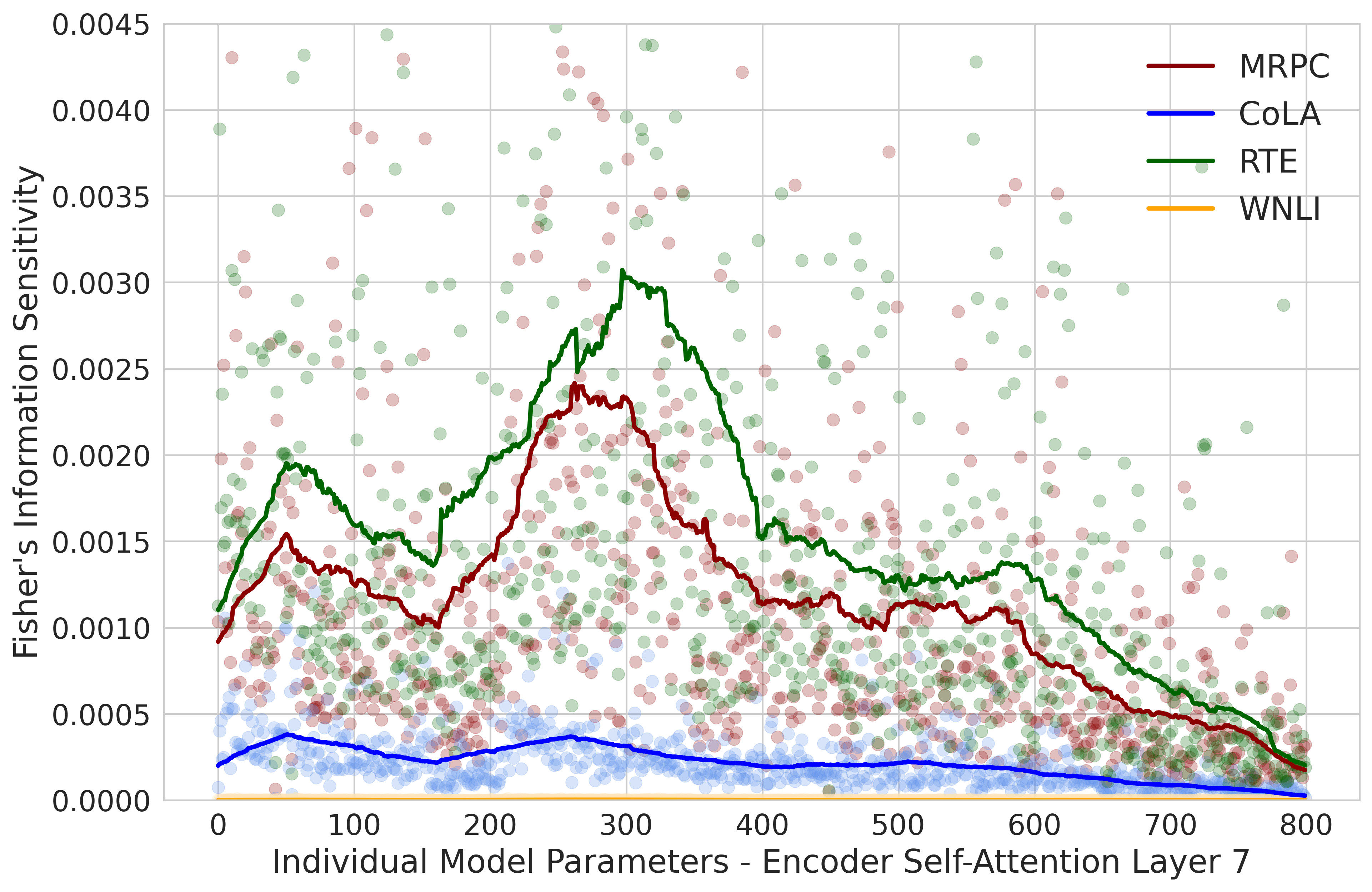}
    \includegraphics[scale=0.22]{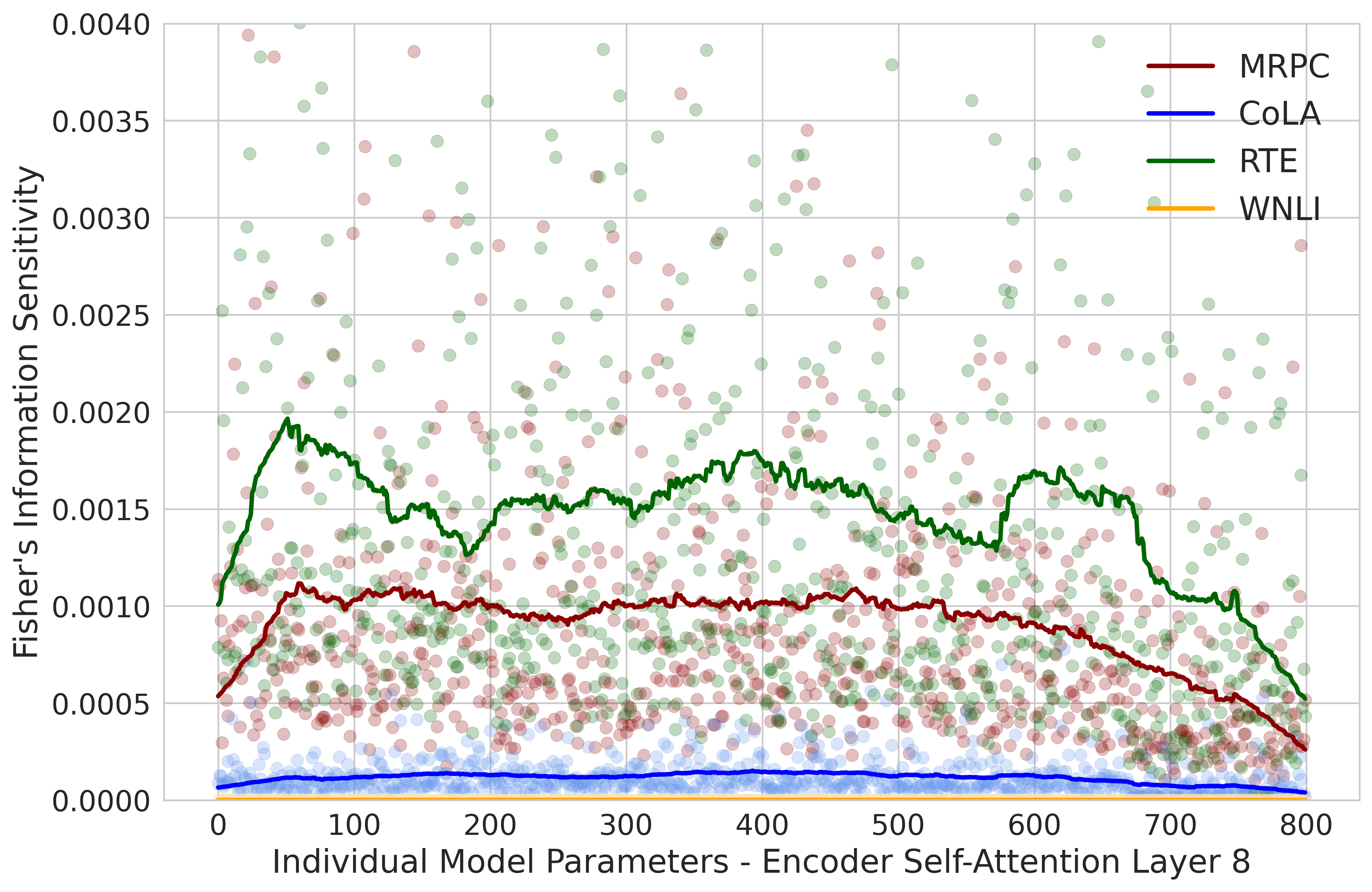}
    \includegraphics[scale=0.22]{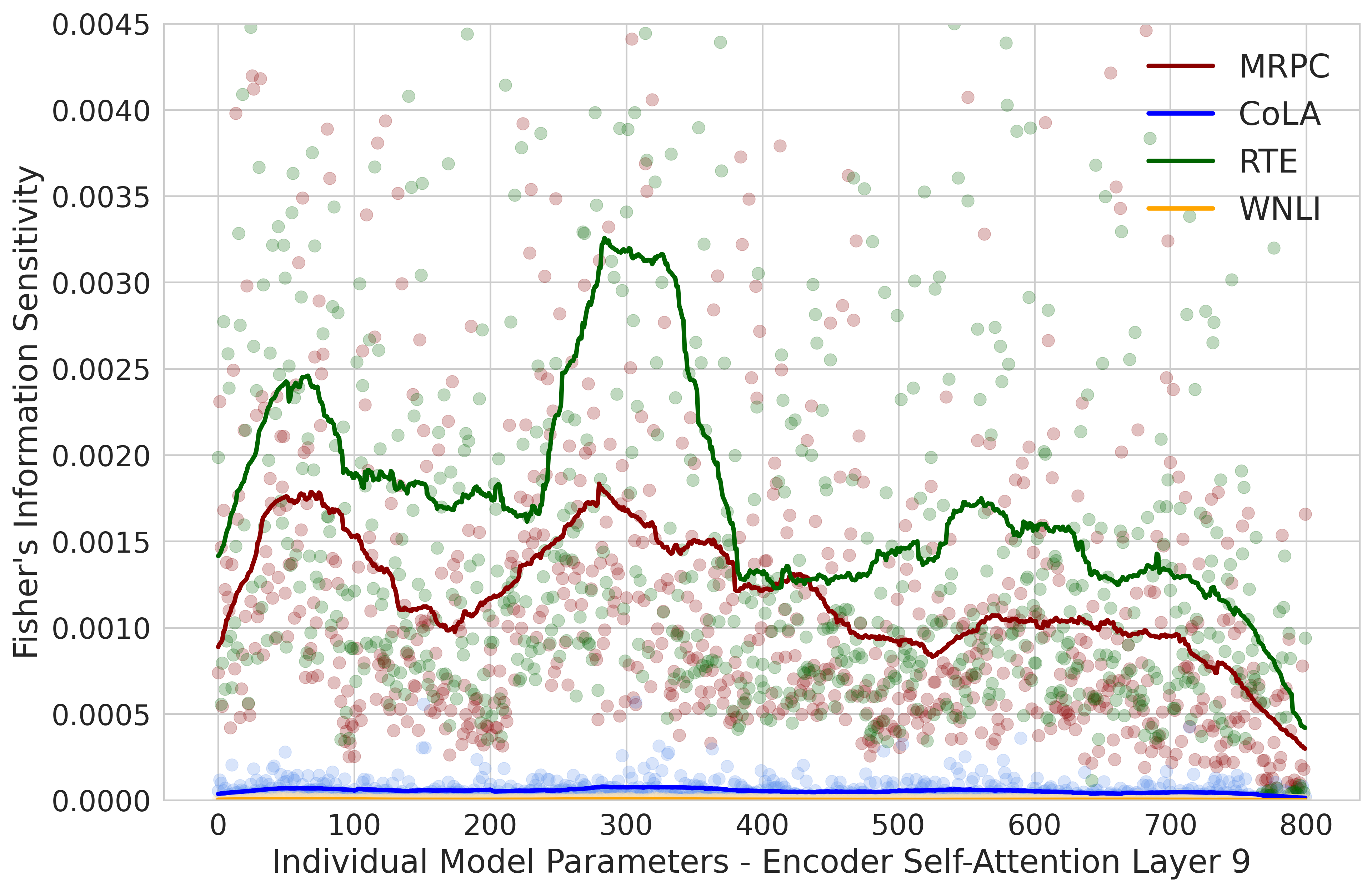}
    \includegraphics[scale=0.22]{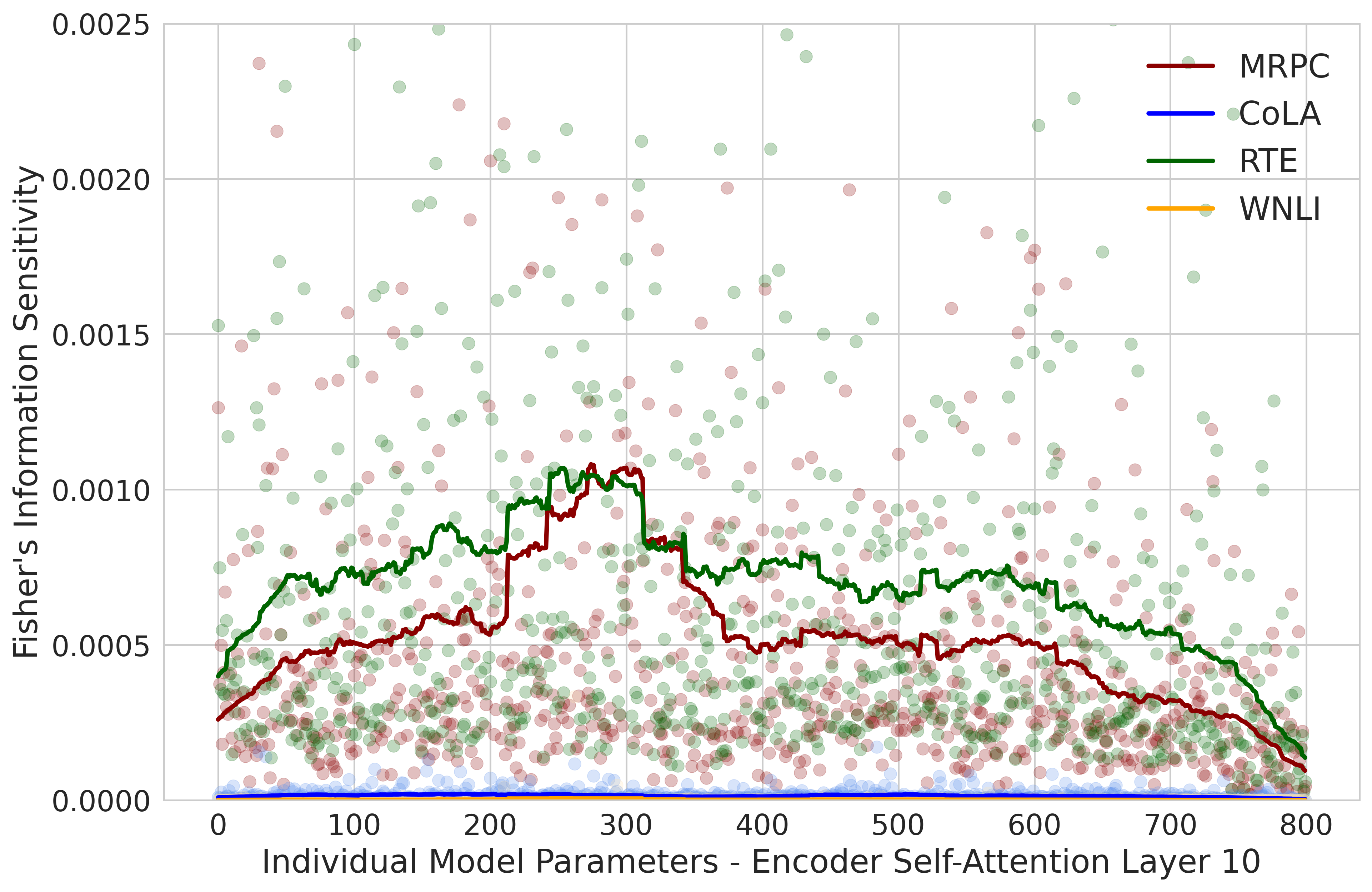}
    \includegraphics[scale=0.22]{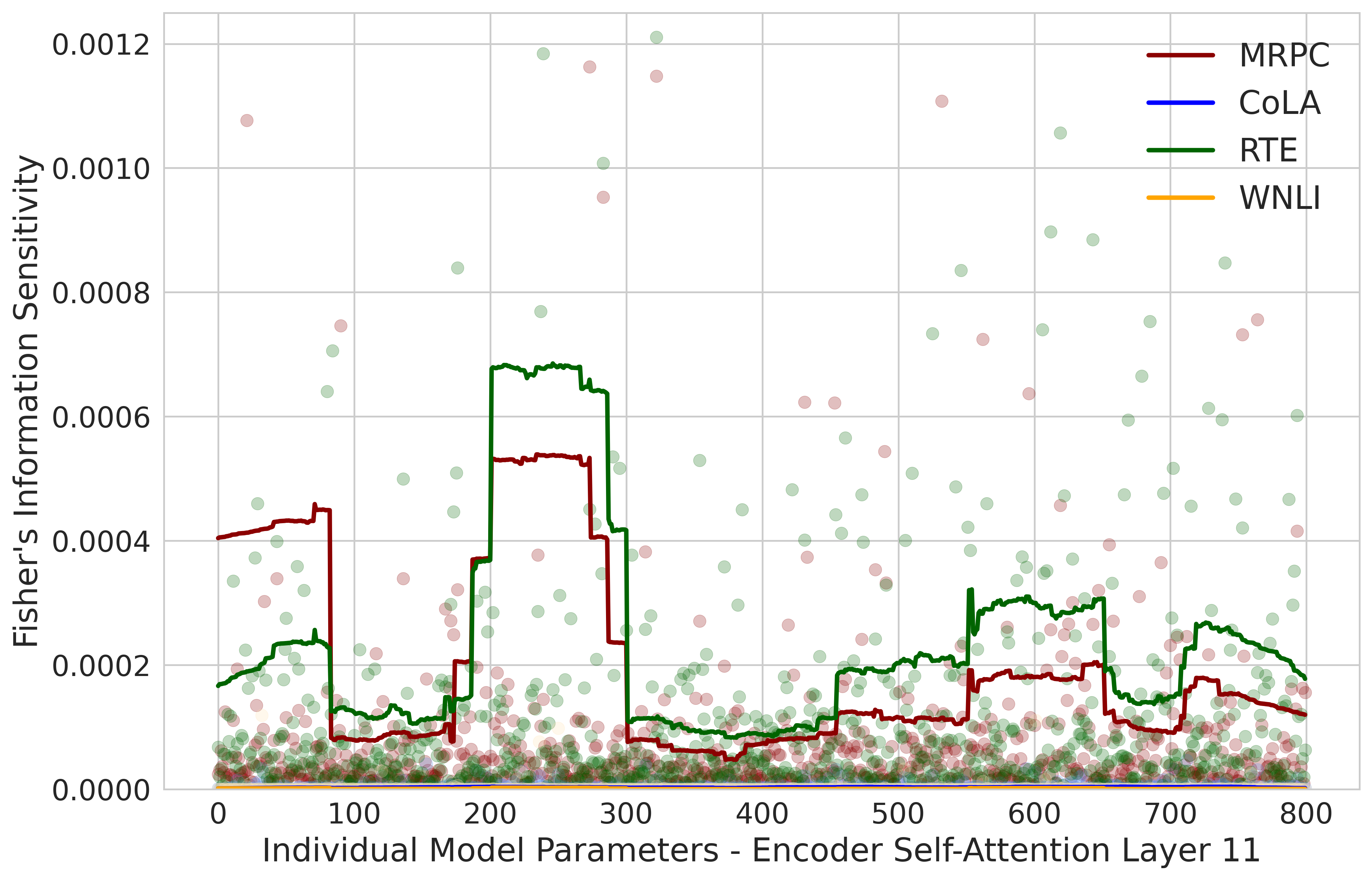}
    \includegraphics[scale=0.22]{images/fisher/12.png}
    \caption{Given Fisher parameter sensitivities $I(\theta)$ the self-attention encoder layers for four different models based on continued training of the base BERT model on four datasets: $I_{arith}(\theta)$ on an arithmetic reasoning and $I_{CoLA}(\theta)$, $I_{MRPC}(\theta)$, $I_{RTE}(\theta)$ on GLUE tasks CoLA, MRPC, and RTE respectively, this plot takes the $n=800$ most crucial parameters based on $I_{arith}(\theta)$ and showcases how sensitive those \textit{same} parameters are to the GLUE tasks based on $I_{CoLA}(\theta)$, $I_{MRPC}(\theta)$, and $I_{RTE}(\theta)$.}%
    \label{fig:allfishers}
\end{figure*}

In \textcolor{blue}{\textsection 1.2}, we substantiated the linguistic forgetting of LLMs through parameter sharing tendencies of the model with illustrations of the parameter sensitivities across different tasks for the first (1st) and last (12th) self-attention encoder layer of the transformer. Here, through Figure \ref{fig:allfishers}, we show that the findings hold across all self-attention encoder layers of model. Further, it is interesting to observe that the task CoLA shares more parameters with the Arithmetic task in the earlier layers compared to the latter layers.

\subsection{Designing {\model}}
\subsubsection{Hyperparameterization for $\mathcal{L}_{REG}$}
The intuition for the selection of hyperparameter $\lambda_{1}$ within the range \{$1e^{-3},1e^{-4}$\} was to scale-match the exceedingly large values of regression loss $\mathcal{L}_{REG}$ to the cross-entropy loss $\mathcal{L}_{CE}$ during the intial phases of training where incorrect predictions of target numerals are frequent. In addition to evaluating the model convergence with $\lambda_{1}$ set to these constants, we also evaluate the following update schedule configurations for $\lambda_{1}$:

\begin{algorithm}
\caption{Update Schedule 1}
\begin{algorithmic}

\State $\lambda_{prev} \gets 1e^{-4}$
\For{i in epochs}
    \State $\lambda_{current} \gets \frac{\mathcal{L}_{REG}}{\mathcal{L}_{CE} + \mathcal{L}_{REG}}$
    \State $\lambda_{1} \gets 0.99 * \lambda_{prev} + 0.01 * \lambda_{current}$
    \State $\lambda_{prev} \gets \lambda_{1}$
\EndFor
\end{algorithmic}
\end{algorithm}

\begin{algorithm}
\caption{Update Schedule 2}
\begin{algorithmic}

\State $\lambda_{prev} \gets 1e^{-4}$
\For{i in epochs}
    \State $\lambda_{current} \gets \frac{\mathcal{L}_{REG}}{\mathcal{L}_{CE} + \mathcal{L}_{REG}}$
    \State $\lambda_{1} \gets 0.01 * \lambda_{prev} + 0.99 * \lambda_{current}$
    \State $\lambda_{prev} \gets \lambda_{1}$
\EndFor
\end{algorithmic}
\end{algorithm}

\subsubsection{Model Training Configurations}
The models BERT$_{Arith}$, GenBERT, and {\model} all share the base BERT architecture. The baseline GenBERT has been employed as-is with the model that the authors provide used for comparative evaluation. For models BERT$_{Arith}$ and {\model}, these are initialized as pre-trained base BERT models with 160M parameters and further trained on randomly sampled $\frac{n}{4}$th of the arithmetic portion of GenBERT's training data. The pre-trained base BERT model is loaded from the HuggingFace library \citep{huggingface}.

The scheme for training follows BERT's standard training protocol of using masked-language modeling. However, instead of randomly masking 15\% of the tokens as done in BERT, we mask the result of the each sample quantitative prompt. For instance, from Figure \ref{fig:samples}, for the sample  \textit{61176.23 - 46741.95 = 14434.28}, the models BERT$_{Arith}$ and {\model} are trained to predict 14434.28 for the masked prompt \textit{61176.23 - 46741.95 = [MASK]}. With the standard sequence size of 512 for BERT, the models were trained for 60 epochs in a cluster of 4 Tesla P100 GPUs.

\end{document}